\documentclass[journal]{IEEEtran}
\usepackage{cite, amssymb}
\usepackage{makecell}
\usepackage{color, array,multirow}
\usepackage{algpseudocode}% http://ctan.org/pkg/algorithmicx
\usepackage{algorithm, amsmath}% http://ctan.org/pkg/algorithm
\usepackage{courier}

\ifCLASSINFOpdf
   \usepackage[pdftex]{graphicx}
\else
\fi
\begin{document}
\title{Adaptive and Robust Lung Cancer Screening Using \\ Memory-Augmented Recurrent Networks}

\author{Aryan~Mobiny\IEEEauthorrefmark{1}\IEEEauthorrefmark{2},
    ~Supratik~Kumar~Moulik, Hien~Van~Nguyen\IEEEauthorrefmark{1}\\ 
\IEEEauthorrefmark{1}Department of Electrical and Computer Engineering, University of Houston \\ \IEEEauthorrefmark{2} \texttt{Email: amobiny@uh.edu}\\
   }

\maketitle

\begin{abstract}
In this paper, we investigate the effectiveness of deep learning techniques for lung nodule classification in computed tomography scans. Using less than 10,000 training examples, deep networks perform 4.8 times better than a standard radiology software under the cross-error measure. For the first time, we provide a systematic comparison between deep networks and radiologists for CT lung nodule data. This includes asking radiologists to examine important networks' features and perform blind relabeling of networks' mistakes. Visualization of the networks' neurons reveals semantically meaningful features that are consistent with the clinical knowledge and radiologists' perception. Our paper also proposes a novel framework for rapidly adapting deep networks to the radiologists' feedback, or change in the data due to the shift in sensor's resolution or patient population. The classification accuracy of our approach remains above 80\% while popular deep networks' accuracy is around chance. Finally, we propose using inconsistency density function, computed by a recurrent network, as a way to discover potentially noisy labels in lung data. Experimental results show that our method can remove $95\%$ of incorrect labels in a synthetic case, and filter out meaningful lung nodule images for relabeling.

%In this paper, we investigate the effectiveness of deep learning techniques for lung nodule classification in computed tomography scans. Using less than 10,000 training samples, we're able to train a neural network that can perform quite well on the source domain, but fails to generalize to the target domain. Therefore, we propose a novel framework for rapidly adapting deep networks to the radiologists' feedback, or change in the data due to the shift in sensor's resolution or patient population. The classification accuracy of our approach remains above 90\% in target domain while popular deep networks' accuracy is around chance level. Finally, we demonstrate that the order of feeding the samples from the target domain to the recurrent structure matters. We then propose a sorting method according to the information gained from each sample. Our experimental results proves the effectiveness of the proposed method in designing an adaptive model which is capable of perfectly adapt to the changes in lung CT data.

%Deep neural networks has been shown to perform quite well and achieved state-of-the-art performance in medical imaging tasks. However, their adaptability to the shift in data domain is still a matter of concern. Especially in the case of medical images, changes in the subject population, data acquisition equipment and their settings can lead to a major variability or shift in the data.

\end{abstract}


% \fi
\IEEEpeerreviewmaketitle

\section{Introduction}
\IEEEPARstart{L}{ung} cancer is consistently ranked as the leading cause of the cancer-related deaths all around the world in the past several years, accounting for more than one-quarter (26\%) of all cancer-related deaths \cite{siegel2017cancer}.  The stage at which diagnosis is made largely determines the overall prognosis of the patient. The five-year relative survival rate is over 50\% in early-stage disease, while survival rates drop to less than 5\% for late-stage disease \cite{siegel2017cancer}. The main challenge in lung cancer screening is detecting lung nodules \cite{national2011reduced, van2009management}. Radiologist fatigue, increasing image set size, and stringent turn-around-time requirements are just a few of the factors which negatively impact detection rate for lung nodules. Many studies have documented the occurrence of diagnostic errors in clinical practice, caused by many different contributing factors which can generally be divided into person-specific (such as radiologists' complacency etc) and environment-specific issues (e.g. inadequate equipment, staff shortages, excess workload, etc.)  \cite{brady2012discrepancy}, \cite{brady2016error}.

Computer-aided diagnosis (CAD) systems aim to improve the radiologist's performance in terms of diagnostic accuracy and speed \cite{shewaye2016benign}. The role of CAD systems in lung nodule detection and screening has been demonstrated over the years \cite{awai2004pulmonary}, \cite{sahiner2009effect}, as well as their role in distinguishing benign from malignant nodules \cite{shewaye2016benign}, \cite{lee2010random}. However, automated identification of nodules from non-nodules is quite challenging mainly due to the large variation in sizes, shapes, and locations of the nodules \cite{dou2017multilevel}. There are also different categories of nodules (such as solitary, pleural, and ground-glass opacity) contributing to the diversified contextual environment around the nodule tissue \cite{firmino2014computer}.

\begin{figure}[!t]
\centering
\includegraphics{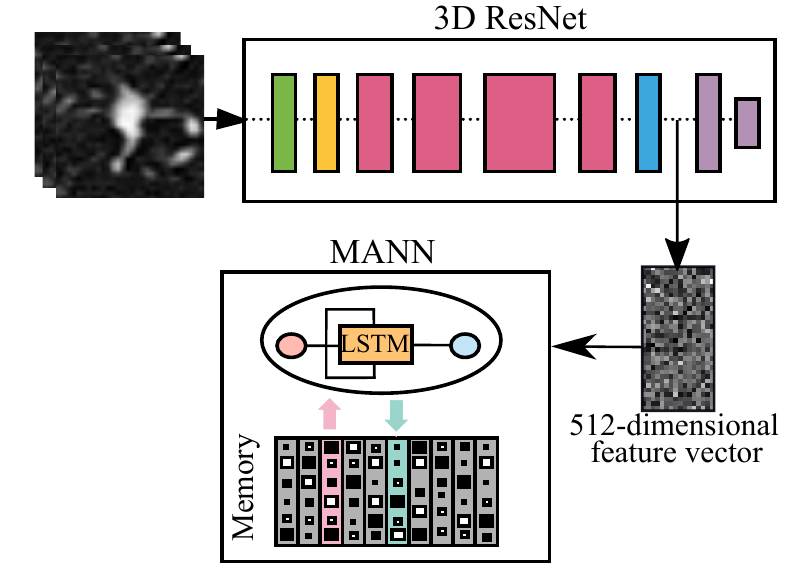}
\caption{Architecture of the adaptive lung nodule classifier, consisting of a 3D ResNet and a memory recurrent network.}
\label{merge_struct}
\vspace{-5mm}
\end{figure}

The performance of a conventional CAD system depends heavily on the intermediate image processing stages (such as extracting hand-crafted morphological and statistical features) which are both time-consuming and subjective \cite{shen2017multi}. In recent years, deep learning technology has attracted considerable interest in the computer vision and machine learning community \cite{lecun2015deep, krizhevsky2012imagenet, girshick2014rich, donahue2015long, jia2014caffe, kalchbrenner2014convolutional, graves2013speech, xu2015show}. Deep neural networks (DNNs) have an advantage of automatically capturing the image's higher level feature representation directly from the input pixel data. This leads to powerful features more tuned to specific tasks of medical image analysis \cite{xu2014deep,liao2013representation, zheng20153d, cheng2016deep, greenspan2016guest, bar2015chest, roth2014new, roth2015deep}. Recent work has explored deep networks for detecting lung pathology \cite{kumar2015lung, cheng2016computer, bar2015chest, mobiny2018fast}. In the context of pulmonary nodule classification in CT images, Hua et al. \cite{hua2015computer} introduced models of a deep belief network and a convolutional neural network that outperforms the conventional hand-crafted features. Setio et al. \cite{setio2016pulmonary} proposed a multiview convolutional network lung nodule detection. There's a lack of systematic analysis on how well deep networks perform compared to radiologists on lung nodule classification task. In addition, while most work in this domain focuses on improving the classification accuracy for a static dataset, the problem of adapting a classifier to changes in lung CT data is largely under-investigated. Finally, the ground truth labels of lung nodules are not absolutely correct. There is an urgent need for techniques capable of removing noisy labels from the training dataset. This paper makes the following contributions: 
\begin{enumerate}
    \item For the first time, we provide a systematic comparison between deep networks and radiologists on lung nodule classification task. The comparison includes asking radiologists to examine important networks' features and perform blind relabeling of networks' mistakes.
    \item We propose a novel framework for rapidly adapting deep networks to the radiologists' feedback using a memory-augmented recurrent network. We demonstrate that our classifier is also robust to shifts in data distribution caused by the variation of sensing technology or patient population.
    \item To mitigate the problem of incorrect labels pertinent to lung nodule data, we propose using inconsistency density function, computed by a recurrent network, as a way to discover potentially noisy labels in lung data. We validate the effectiveness of our algorithm on both synthetic and real datasets. 
\end{enumerate}

The rest of this paper is organized as follows: Section~\ref{radiologist_comparison} provides in-depth analysis on the performance of deep networks versus radiology software and radiologists. Section~\ref{rapid_adaptation} explains how we adapt a classifier to radiologists' feedback or changes in the data. How to remove noisy labels using a recurrent network is the focus of Section~\ref{noise_discovery}. Section~\ref{conclusion} concludes the paper with future research directions. 

%In this paper, we propose a state-of-the-art "expert in the loop" framework using the idea of one-shot learning which enables the radiologist to modify the output of the CAD system (i.e. CNN classifier) by providing only a few feedbacks. The first advantage of this framework over its counterparts is that it allows radiologists to check the errors made by the network in a real-time manner and track its improvement happening by providing sequential feedbacks. This will eventually enable radiologists to calibrate their trust in the CAD system more effectively and stop the process as soon as they are convinced of the accuracy of the results. Second, there is no need to re-train the network from scratch (which is both time-consuming and inefficient) each time we receive new sets of data from relatively different distributions. The final goal of designing such framework is to be able to rapidly understand and adapt to the new CT scan images. This paper makes the following contributions:

\section{Comparison of Deep Networks and Human Radiologist on Lung Nodule Classification}
\label{radiologist_comparison}
This section investigates the effectiveness of two well-known deep networks for lung nodule classification. We compare the results against accuracies of a radiology software and human experts. We also provide visualization of networks' filters to help us better understand the properties of deep networks trained on lung nodule data.
\vspace{-2mm}
\subsection{Dataset}
The study included 226 unique Computed Tomography (CT) Chest scans (with or without contrast) captured by General Electric and Siemens scanners. The data was preprocessed by an automated segmentation software in order to identify structures to the organ level. From within the segmented lung tissue, a set of potential nodule point is generated based on the size and shape of regions within the lung which exceeds the air Hounsfield Unit (HU) threshold. Additional filters, based on symmetry and other common morphological characteristics, are applied to decrease the false positive rate while maintaining very high sensitivity.

\begin{figure}[!t]
\centering
\includegraphics[width=0.5\textwidth]{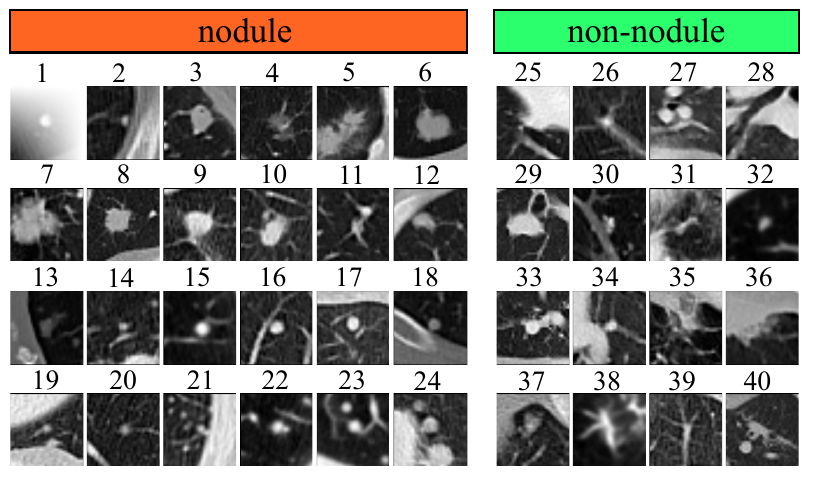}
\caption{Sample images of nodules (Left) and non-nodules (Right). Each image is a slice along 2D axial plane in the middle of the volume.}
\label{sample}
\vspace{-5mm}
\end{figure}

Bounding boxes with at least 8 voxels padding surrounding the candidate nodules are cropped and resized to $32\times 32\times 32$ pixels. Each generated candidate is reviewed and annotated by at least one \emph{board certified radiologist}. From all the generated images (about 7400 images), around 56\% were labeled as nodules and the rest non-nodules. Figure \ref{sample} shows examples of extracted candidates and the corresponding labels provided by radiologists. These images illustrate the highly challenging task of distinguishing nodules from non-nodule lesions. One reason is that the pulmonary nodules come with large variations in shapes, sizes, types, etc. In Figure \ref{sample}, examples of solitary (1), sub-pleural (2), cavitary (3) and ground-glass (4) nodules are depicted. (5) is a more complicated sample containing a mixed solid and ground-glass nodule with irregular margins. While nodules are commonly known as spherical lesions, they also often have a non-spherical shape (12-13) and irregular margins. These irregularities can be caused by vessels and/or spiculations (6-11). Other objects and tissues might also appear in the nodule samples, such as single or multiple blood vessels (14-17), chest wall (18-19), lung recess (19), etc. Moreover, one nodule image can also contain several nodules of different shapes and sizes (21-24). 

 The second reason which hinders the identification process is the non-nodule candidates mimicking the morphological appearance of the real pulmonary nodules. Examples are calcification (30), short vessels (31-34), scarring (35), infection (36-37), vessels with motion artifact mimicking a ground-glass nodule (38), septical thickening (39). Some images might also contain a nodule, but is centered on another tissue (such as a vessel in (40)) and so is labeled as non-nodule. For all these reasons, the detection and classification of lung nodules is a challenging task, even for experienced radiologists.

\subsection{Deep Network Architectures} 
We modify two  well-known deep network architectures-- AlexNet \cite{krizhevsky2012imagenet} and ResNet \cite{he2016deep} to make them compatible with 3D data. The modified AlexNet contains eight layers (five convolutional layers followed by three fully-connected layers) similar to the original AlexNet. The local response normalization layers after the first, second, and last convolutional layers proposed in the original paper are replaced by batch normalization (BN) layers \cite{ioffe2015batch}. We also add BN before all other ReLU layers. Empirically, BN enables much faster convergence rate during training. It also has regularization effect like dropout because of computing the statistics on every mini-batch (rather than the entire training examples) \cite{ioffe2015batch}. We use a smaller stride of 1 and filter size of 4 for the first convolutional layer. This results in to more distinctive features and fewer dead filters, as demonstrated in \cite{zeiler2014visualizing}. The dropout rate is set to 50\% \cite{krizhevsky2012imagenet} in the first two fully-connected layers to prevent over-fitting.

We use 50-layer ResNet similar to what is described in \cite{he2016deep} with some modifications applied to fit our data dimensions. At the top of the network, we use a convolutional layer including 32 filters of size $4\times4$ with stride 2, followed by a $2\times2$ max-pooling layer with stride 2. These are followed by four so-called bottleneck blocks which are described in details in \cite{he2016deep}. Each of the three convolutional layers in a single bottleneck block follows by BN which is applied before the ReLU nonlinearity. A fully-connected layer with 50 hidden units is also added to the network and before the classification layer. This layer helps to improve our classification results. Similar to AlexNet, a dropout with 50\% rate is used in this fully-connected layer. We also apply 2D versions of these networks to 2D slices of lung nodule volumes along the x-axis. We choose x-axis because it contains most information according to radiologists' feedback. Having both 2D and 3D images enable us to compare to quantify the contribution of the third dimension. 

\vspace{2mm}
\noindent \textbf{Optimization Settings:} For both networks, we train using ADAM optimizer \cite{kingma2014adam} and cross-entropy loss function. We perform data augmentation by randomly rotate nodule volumes around the center for 2D images and along all three axes for 3D images. We set the maximum rotation degree to $90^o$ and $45^o$ for 2D and 3D networks respectively so as to prevent introducing too much distortion to the images. Afterwards, pepper noise was added to the rotated images. It's similar to applying dropout to the visible layer, i.e., the input. The dropout rate is set to 5\%.

\subsection{Comparison of Deep Networks and Radiology Software} We first compare the performances of deep networks against an automated radiology software. The automated image segmentation software has an algorithmic method for separating true from false candidate nodule points.  By utilizing the lung segmentation data and the voxel density value, the center of each candidate point is estimated as the point which is the most equidistant from surrounding air density (-200 HU) lung points.  Radial density analysis is subsequently performed to determine the symmetry characteristic of the candidate point.  Specifically, the nodule is categorized as either mostly spherical or mostly cylindrical.  This geometric analysis forms the backbone of the automated nodule analysis.  

%There are several key limitations to the method which necessitated the addition of machine learning algorithms.  First, for practical computation reasons, only limited number of radial direction are analyzed which means that a vessel had to be oriented along the path of the angular steps in order to be determined as such.  The symmetry analysis also requires the establishment of an edge threshold which becomes problematic is the setting of inhomogeneous or ground glass nodules.  In addition, if a nodule is situation along a blood vessel, the nodule is likely to be mischaracterized as a vessel; given that most metastatic disease to the lungs is spread hematogenously, this is the most clinically significant limitation.  Finally, the variations in morphology, homogeneity and overall size contribute to the limitation of the algorithm.  

\begin{figure}[!t]
\centering
\includegraphics[width=0.4\textwidth]{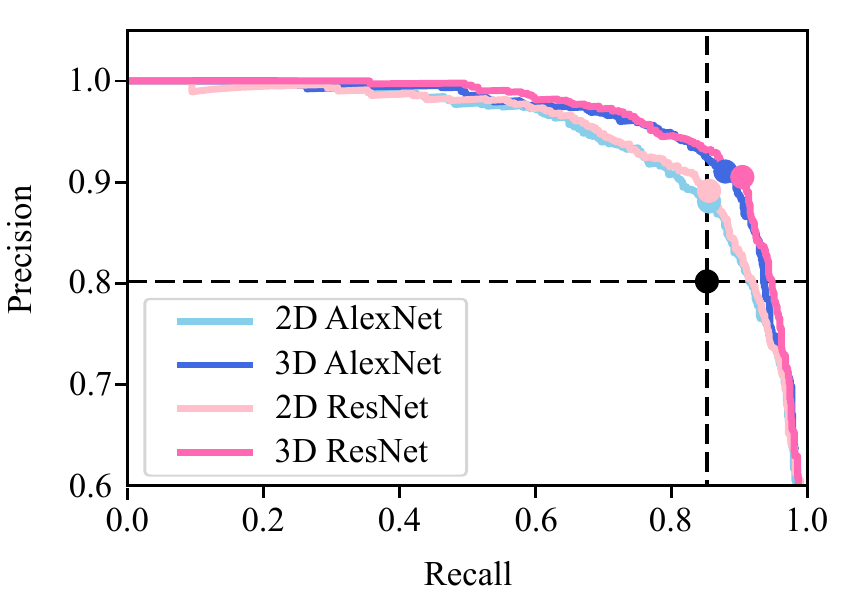}
\caption{Precision-recall Curves for the Designed Networks. The points on the curves shows the precision and recall values at the threshold of 0.5. The black dot at the intersection of dotted lines depicts the precision and recall values for the automated software}
\label{pr_curve}
\end{figure}

For the binary classification of the unbalanced classes, performance is quantitatively determined via the precision, recall (sensitivity), specificity and error rate metrics. Validation results for the 2D and 3D CNNs, for both AlexNet and ResNet, are provided in Figure \ref{pr_curve}. The specified point on each of the curves shows the precision and recall values at the default threshold of the classifier (i.e., 0.5). The black point shows the precision-recall of the automated software. Table I includes the performance values of all CNNs, as well as those of the automated software (which is treated as the baseline result). To get a better comparison, the results for the automated software was also computed and presented over the same validation set. The best result of each column is shown in bold.

\begin{table}[b]
% \resizebox{0.85\textwidth}{!}
\label{classification_result}
\caption{Comparison of the Deep Networks and Automated Radiology Software.}
\begin{tabular}{ccccc}
\hline
& \textbf{Presicion} & \textbf{Recall} & \textbf{Specificity} & \multicolumn{1}{l}{\textbf{Error rate}} \\ \hline
\thead{Automated Software \\ on all images} & 80.19\% & 85.21\% & 71.78\% & 21.79\%    \\
\thead{Automated Software \\ on validation set} & 73.65\% & 82.13\% & 70.01\% & 25.34\%    \\
\hline
2D AlexNet     & 88.09\% & 85.52\% & 88.32\% & 13.09\%  \\
2D ResNet      & 89.14\% & 85.52\% & 89.47\% & 12.51\%     \\  
3D AlexNet     & \textbf{91.05}\% & 87.92\%          & \textbf{91.26}\%    & 10.42\%   \\           
3D ResNet  & 90.51\% & \textbf{90.42}\% & 90.42\% & \textbf{9.58}\%  \\ 
\hline
\end{tabular}
\end{table}

As shown in Table I, all CNN networks outperform the automated software by a large margin. Moreover, 3D networks improve over their 2D counterparts by 3\% error points. This shows that the 3D networks are capable of encoding and exploiting the complicated anatomical surrounding environments of the volumetric image. 3D ResNet achieves the lowest error rate of 9.58\%, outperforming 3D AlexNet with the error rate of 10.42\%. 
%ResNet also results in a higher sensitivity (recall of 90.42\% compared with 87.92\% of AlexNet) meaning that it misses less nodules. However, 3D AlexNet gives a slightly higher precision and specificity, reporting fewer false positives. 
However, looking at the precision-recall curve in Figure \ref{pr_curve}, there's no strong evidence for us to prefer one network over the other for lung nodule classification task.

We compared the performance of the automated algorithm and the deep learning approach by computing the number of examples that are correctly classified or misclassified by either of them. These values are presented in terms of percentages in Table \ref{cross_corr}. We compute the \emph{cross-error measure}, defined as the ratio of errors made by one method, and not by the other. Table \ref{cross_corr} indicates that the designed ResNet is performing 4.8 times (19.84\% of ResNet compared with 4.08\% of automated software) better than the standard automated radiology software under this measure.

\begin{table}[!t]
\renewcommand{\arraystretch}{1.3}
\caption{Cross-comparison of the classification methods}
\label{cross_corr}
\centering
\begin{tabular}{r|l}
\textbf{Both Correct}     & \textbf{Software correct Only} \\
72.46\%                    & 4.08\%                   \\ \hline
19.84\%                    & 5.50\%                   \\
\textbf{ResNet Correct Only} & \textbf{Both Wrong}    
\end{tabular}
\end{table}

\subsection{How Deep Networks Fare Against Radiologists?} \label{deepnet_vs_radiologist}
\noindent \textbf{Misclassified images are difficult for radiologists:} Typically, radiologists have an error rate in range of 2-20\%. \cite{brady2012discrepancy}. So it could be possible that the images misclassified by the network are actually the correct classification and the ground truth could be wrong. In order to verify this, we asked a board-certified radiologist to relabel all of the images misclassified by the 3D-ResNet and see if the new labels match with the ground truth. 125 images were provided to the radiologist for relabeling, without him knowing how they were labeled in the first time. The number of matched labels was 36 out of 115 (31.3\%) and the number of mismatched labels were 79 out of 115 (68.69\%). The radiologist was not able to decide on the label for the remaining 10 images. It is surprising that the mismatch rate is even higher than 50\%, which is the expected mismatch rate if the radiologist was the act randomly. This demonstrates that samples classified incorrectly by deep networks are also highly confusing for the radiologist. The most common cases associated with mismatched labels are infection (17 out of 115), granuloma (4 out of 115), scarring (9 out of 115), blood vessel (6 out of 115) and atelectasis (10 out of 115). These constitute 58.2\% of the mislabeled images. 

\begin{figure}[h]
\centering
\includegraphics[width=0.5\textwidth]{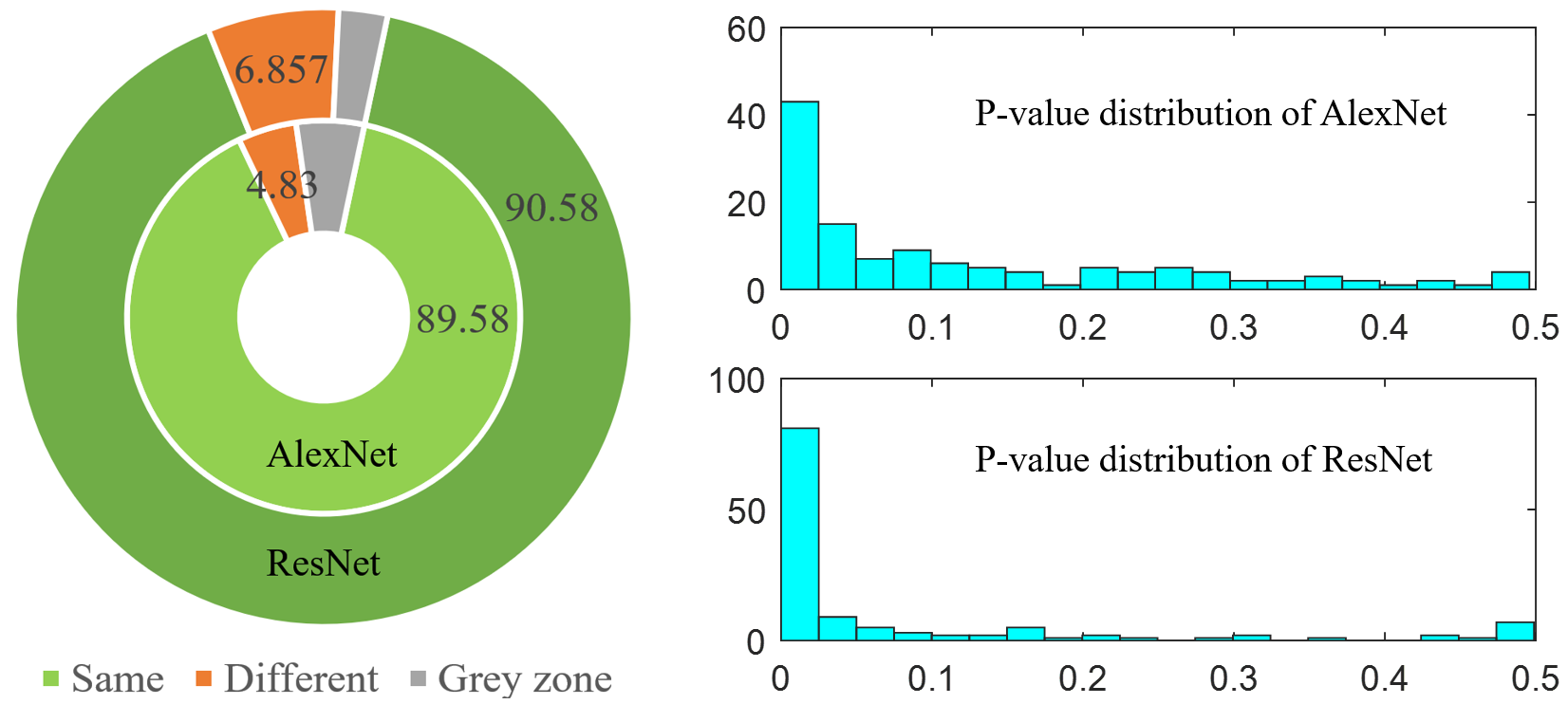}
  \caption{Comparison of deep networks and radiologist classification (left), and distribution of p-values (right).}
\label{rad_clas_compare}
\end{figure}

\vspace{2mm}
\noindent \textbf{How much do deep networks differ from radiologists?} We begin with specifying the null hypothesis as $P_\text{network}(y | \mathbf{x}) = P_{\text{radiologist}}(y | \mathbf{x})$. For a given sample, we assume the network picks a label randomly under Bernoulli distribution given by $P_\text{network}(y | \mathbf{x})$. This allows us to compute the p-value given observations of labels provided by radiologists under the null hypothesis. For example, if a radiologist diagnoses that the sample is nodule the first time, and non-nodule the second time, p-value is equal to $P(\text{one nodule in two trials} | P_\text{network})$. Applying this estimate over all the relabeled samples give us 125 p-values. We also compute \emph{q-value} \cite{storey2003statistical} from the distribution of p-value to estimate the false detection rate for setting certain p-value (and those below it) to be significant. We reject the null hypothesis, or decide radiologist and classifier decisions are significantly different, when both p-value and q-value are less than 0.05. Figure~\ref{rad_clas_compare} shows that $6.85\%$ and $4.83\%$ error points are significant, whereas $2.56\%$ and $5.59\%$ fall into grey zone for ResNet and AlexNet, respectively. The reason that ResNet has higher percentage of significant samples is because its prediction scores are often more extreme than AlexNet. In conclusion, deep networks are at least $4.83\% \rightarrow 6.85\%$ different from radiologists prediction. However, the fact that more than 50\% of samples misclassified by deep networks have opposite labels during relabeling makes it uncertain to conclude whether the networks or radiologists did better.

%We use \emph{p-value} to estimate how many samples where radiologists' decisions are significantly different from deep networks. Conditional distribution of label $P(y |  \mathbf{x})$ varies from image to image. This gives rise to multiple-hypothesis testing problem, where each image constitutes one hypothesis. We compute \emph{q-value} \cite{storey2003statistical} from the distribution of p-value to estimate the false detection rate for setting certain p-value (and those below it) to be significant. 
%to provide confidence of our estimate. The reason is that low p-values can come from both null hypothesis and alternative hypothesis. This is because the distribution of p-values under the null hypothesis is uniform \cite{storey2003statistical}. It is therefore not straight forward to estimate the fraction of classifier's mistakes that is truly different from radiologist's diagnoses. q-value provides a systematic way to estimate the false detection rate if we set certain p-value (and those below it) to be significant.

\subsection{Visualization}
To visualize what features deep networks capture, we use the visualization technique introduced in \cite{zhou2014object}. Since some images have a very complicated appearance, it's hard to tell what is the object type that is detected by a specific unit. For this reason, in addition to real lung data,
 we ask radiologists to help generate synthetic data representing common patterns of nodule and non-nodule lesions. Looking at the activation regions in synthetic images with simpler appearance can help to better understand the exact representation learned by each filter. Synthetic data contains spherical shapes mimicking the nodule examples and tube shapes similar to non-nodule cases, especially vessels. 

\begin{figure}[h]
  \begin{center}
    \includegraphics[width=0.48\textwidth]{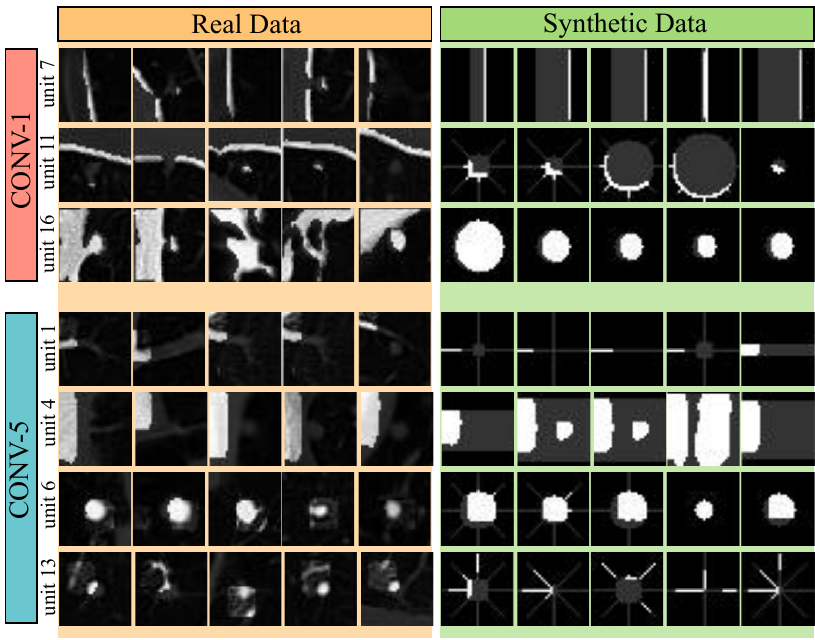}
  \end{center}
  \caption{Examples of object detector units from the first and last convolution layers of the AlexNet. The unshaded area in each image depicts the activation region. Each row contains the five maximally activating images. Results are presented for both real (Left) and synthetic (Right) images and the same units.}
  \label{visual}
\end{figure}

Figure~\ref{visual} plots the maximally activated images of different filters for AlexNet. The network is trained by the real images, and RF-based segmentation is done for both real (left panel) and synthetic (right panel) data. In each image, the area that is not shaded in gray depicts the activation region. Comparing the activation regions of the units of con1 and conv5 shows that the activation regions become more semantically meaningful with increasing the depth of layers. Units at the early layers are more responsive to simple shapes and edges. For example, unit 7 and 11 of conv1 are responding to edges created by changes in contrast (i.e. change in color) in different directions. Unit 7 captures the edge of vertical chest walls in real data. Since there is no chest wall in synthetic data (and given that this network is trained on the real data), this unit captures the right edge of the vertical tubes which show the same texture and contrast patterns as a chest wall. The same behavior is observed by unit 11 which gets activated when there is a change in color from top to bottom and from white to black. Unit 16 is responding to the whole white area presented in the image.

Unit 1 in conv5 is maximally responding to thin horizontal tube shapes which are mostly vessels. Unit 4 captures the chest wall on the left side of the ROI. Unit 6 is activated by simple nodules placed at the center of the ROI. Unit 13 shows a complicated yet interesting behavior. It's searching the environment surrounding real nodules and responds to the objects such as nodule speculations or small vessels. Activation regions in synthetic images shed more light on the behavior of this unit. 

\section{Rapid Adaptation of Lung Nodule Classifier}
\label{rapid_adaptation}
Radiologist's \textit{trust} in the computer-aided diagnosis (CAD) systems is a key factor in increasing the adoption rate of CAD to clinical practice. Since radiologists are responsible for the final diagnosis, it is important for them to be able to tell CAD systems to correct any detection they think wrong. As a result, the system has to be able to rapidly adapt to radiologists' feedback and further refine its decision. Such interaction is crucial for radiologists to understand how CAD systems work. This in turns will help improve efficacy and efficiency of the radiologist-CAD team \cite{jorritsma2015improving}.  

Another important reason for making lung nodule classifier adaptive is the perpetual problem of domain shift. That is the problem when training data are different from test data. For lung nodule, difference in CT scanner's sensing technology, reconstruction algorithms, or scanning protocol are common reasons causing discrepancy between training and test data. Variation in the patient population is another factor contributing toward shift in data distribution. It has been shown that the classification accuracy reduces dramatically when training and test data come from different distributions \cite{saenko2010adapting, kulis2011you, gopalan2011domain, gong2012geodesic}. Re-training a lung nodule classifier is both expensive and time-consuming, which can disrupt the clinical workflow. 

In this section, we propose a framework for adapting lung nodule classifier, using only few feedback, to never-seen-before data distribution. The proposed approach enable radiologists to review a few errors made by a deep network and incorporate his knowledge to correct them. Our classifier use these feedback to further refine its decision.

\subsection{Background on Memory Augmented Recurrent Network} 
Our approach uses a memory augmented neural network (MANN) \cite{santoro2016one}, a variant of Neural Turing Machine \cite{graves2014neural}, as the main building block for processing sequential feedback information. MANN consists of two main components: a) a LSTM network as the main controller, and b) an external memory bank interacting with the main controller through read and write operations. The  external memory is denoted by a matrix $\mathbf{M}_t \in \mathbb{R}^{k\times q}$ where $k$ is the number of memory slots and $q$ is the size of each slot. The model has an LSTM controller that reads and writes to the external memory at every time step. The reading operation is done by a weighted linear combination of all memory slots in external memory as follows:
\begin{equation}
\label{read}
\mathbf{r}_t = (\mathbf{M}_t)^T . \mathbf{w}_t^r
\end{equation}

Here, $\mathbf{r}_t$ is the content vector, and $\mathbf{w}_t^r \in \mathbb{R}^{k\times 1}$ is the read weights computed from the hidden nodes $\mathbf{h}_t$ in LSTM main controller. To write into the memory, Santoro et al. \cite{santoro2016one} designed a module called Least Recently Used Access (LRUA) to access the least used memory locations by computing the \textit{least-used} weights vector, $\mathbf{w}_t^{lu}$, at each time step. The write weights $\mathbf{w}_t^w \in \mathbb{R}^{1\times k}$ is then computed as:

\begin{equation}
\label{write_weight}
\mathbf{w}_t^w \leftarrow \sigma(\alpha)\mathbf{w}_{t-1}^r + (1-\sigma(\alpha)) \mathbf{w}_{t-1}^{lu}
\end{equation}

\noindent
where $\sigma(.)$ is the sigmoid function. Let $i$ be the index of the non-zero element in the one-hot vector $\mathbf{w}_t^w$, then the controller writes in the memory as:

\begin{equation}
\label{write}
\mathbf{M}_t(i) \leftarrow \mathbf{M}_{t-1}(i) + w_t^w(i) \mathbf{a}_t
\end{equation}

\noindent
where $\mathbf{a}_t$ is the linear projection of the current hidden state passed through a \textit{tanh} nonlinearity. The introduction of an external memory enables the recurrent network to store and retrieve much longer-term information compared to LSTM. This frees up the main controller and increases its capacity of learning highly complicated patterns within the data.

\subsection{Adaptive Lung Nodule Classifier via Memory Augmented Recurrent Network}
Suppose $D={\{(\mathbf{x}_i, y_i)\}}_{i=1}^N$ is the initial set of training data. Let $F={\{ \mathbf{x}_j, y_j\}}_{j=1}^M$ denote the feedback samples provided by physicians. Given a sample $\mathbf{x}$, our goal is to estimate its true label by conditioning on the initial training set and the feedback data. We can write the conditional probability as follows:

\begin{equation}
\label{prob}
P(y|\mathbf{x}, D, F) = \frac{P(\mathbf{x}, y, D, F)}{P(\mathbf{x}, D, F)} \nonumber =
  P(y|\mathbf{x}, D) \frac{P(F|\mathbf{x}, y,  D)}{P(F|\mathbf{x}, D)} 
\end{equation}

We observe that the first term $P(y|\mathbf{x}, D)$ only depends on the initial training set and the given input, but not the feedback $F$. For this reason, we train a ResNet to approximate this term. The denominator in the second term does not affect the classifier's decision as it is the same for every $y$. Therefore, we only need to estimate $P(F|\mathbf{x}, y, D)$ to update the classifier's decision. To this end, we model this likelihood function using a MANN. We then merge together the output of MANN and ResNet to form an adaptive system capable of generalizing to new set of nodule images. In practice, we observe that the MANN converges rather slowly. This could be because the recurrent network cannot scale very well to the large number of pixels of input images (in our case, $32\times 32 \times 32 = 32768$ pixels). We mitigate this issue by passing images to a ResNet before feeding them into MANN. Specifically, we use 512-dimensional output of ResNet's average-pooling layer as the features to MANN. Our experimental results show that this modification, denoted as ResNet-MANN, dramatically improves the convergence speed of MANN.

\vspace{2mm}
\noindent \textbf{Training MANN:} To train the recurrent network, we sequentially present of images and their one-step delayed labels  $\{\mathbf{x}_t, y_{t-1}\}_{t=1}^T$. Each sequence is called an \emph{episode}. This simulates the sequential feedback from radiologists during the evaluation phase. The network's parameters are optimized through maximizing the cross entropy between predicted probabilistic scores and the ground truth labels. We train MANN using ADAM optimizer with the same configuration as CNN and minibatch size is set to 16. A grid search is performed to find the best parameter values. The best validation results are achieved using 128 memory slots of size 40 and LSTM controller of size 200.

%More importantly, correct labels are randomly shuffled from episode to episode. For example, nodules are labeled as 1 in some episode (so that non-nodule class is labeled as 0) while their labels are randomly changed to 0 in other episodes. This strategy will finally prevent the MANN from learning to simply mapping the samples to their fixed class labels \cite{santoro2016one}. 

%The goal is to model the conditional probability distribution of $P({y}| \mathbf{x}, S)$ where $S=\{ (\mathbf{x}_i, y_i)\}_{i=1}^m$ is the small set of $m$ examples of input-label-pairs, $\mathbf{x}$ is the given test example, and ${y}$ the corresponding class label. In the deep learning framework, $P$ is parametrized by a neural network trained to perform the mapping. We propose to use a variant of Neural Turing Machine \cite{santoro2016one} to make our classifier adaptive to data shift using only few feedback from radiologists.

\begin{figure}[!t]
\centering
\includegraphics[width=0.5\textwidth]{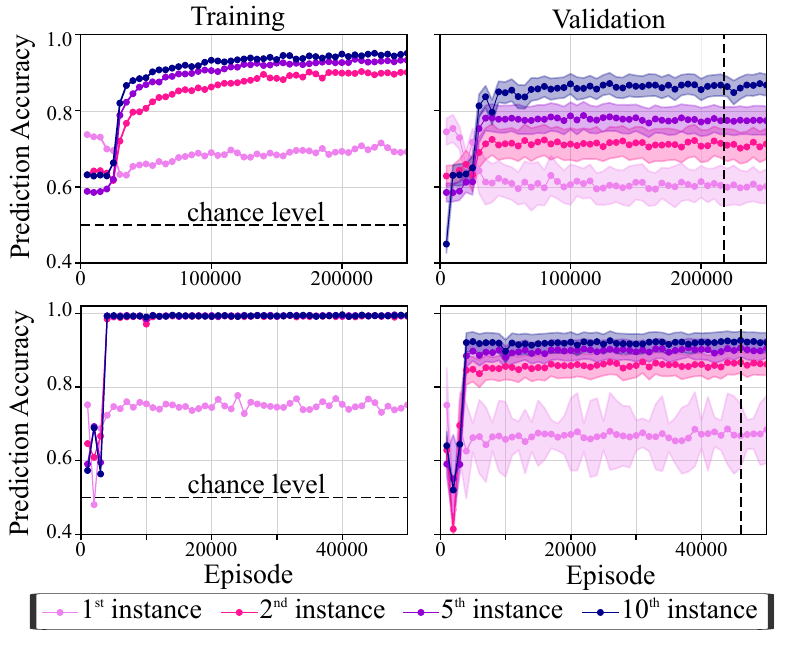}
\caption{Training (Left) and validation-set (Right) accuracies of nodule/ non-nodule classification using MANN and ResNet-MANN. In the right panel and at each specific episode, the validation-set accuracy is presented for a network that is trained for that many episodes. The accuracy was computed and presented as the average accuracy ($\pm$std) over 500 sequence of images (of length 20) selected randomly from the whole validation set. The vertical dashed line in the right panel depicts the results at the episode with the best validation accuracy}
\label{mann_train}
\end{figure}

\begin{figure}[!t]
\centering
\includegraphics{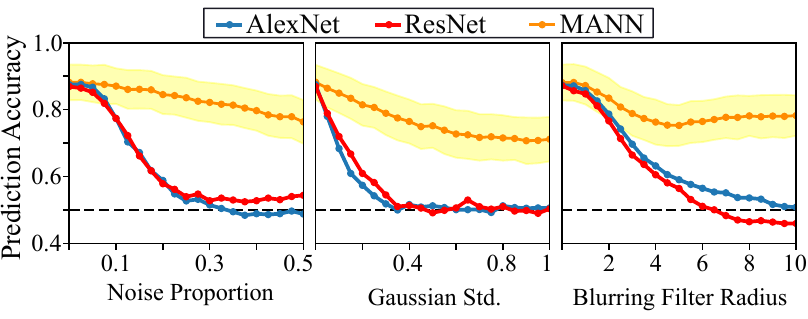}
\caption{Comparison of the validation-set classification accuracies of AlexNet, ResNet, and MANN in response to applying different type of distortions on images; namely, salt and pepper noise (Left), Gaussian noise (Middle), and blurring (Right). The yellow shaded areas depicts the standard deviation of MANN's accuracy in the 500 conducted episodes}
\label{mann_noise}
\vspace{-3mm}
\end{figure}
\subsection{Evaluation of Adaptive Classifier}
\noindent \textbf{Adaptation to feedback:} We first verify the effectiveness of MANN by computing its classification accuracy after a number of feedback. Each feedback contains an image and its label. The training and validation accuracies for the MANN network are provided in the left and right panels of Figure~\ref{mann_train}, respectively. The accuracies are computed for up to 10 feed-backs. For example, the first instance accuracy is the classification accuracy for just the first presentation of samples of each class. The second instance accuracy is the classification accuracy of the second observation of both classes, and so on.

\begin{table}[!t]
\centering
\caption{Validation classification accuracies for MANN and ResNet-MANN}
\label{mann_table}
\resizebox{\columnwidth}{!}{
\begin{tabular}{l|llllll}
\hline
 & \multicolumn{6}{c}{Instance (\% correct)} \\
 \multicolumn{1}{c|}{Model} & $1^{st}$ & $2^{nd}$ & $3^{rd}$  & $4^{th}$ & $5^{th}$ & $10^{th}$ \\ \hline
 \multicolumn{1}{c|}{MANN} & 60.58 & 72.04 & 75.59 & 77.24 & 78.78 & 87.75 \\
 ResNet+MANN& \textbf{66.78} & \textbf{86.70} & \textbf{88.78} & \textbf{90.00} & \textbf{90.60} & \textbf{92.68} \\ \hline
\end{tabular}
}
\end{table}

The average validation accuracies ($\pm$ std) are depicted in the right panels of Figure~\ref{mann_train}. For both architectures, the first instance accuracy is above chance level which indicates that the networks perform \textit{educated guess} for new data samples based on the images it has already seen and stored in the memory. MANN achieves the highest validation accuracy after 226,000 episodes with 60.58\% and 72.04\% for the first and second feed-backs, reaching up to 78.78\% and 87.75\% by the fifth and tenth, respectively. In contrast, ResNet-MANN reaches the highest validation accuracy after only 46,000 episode with 66.78\% and 86.70\% for the first and second feed-backs, reaching up to 90.60\% and 92.68\% by the fifth and tenth, respectively. ResNet-MANN significantly outperforms MANN both in terms of convergence rate and accuracy in adaptive classification setting. Table \ref{mann_table} summarizes the validation accuracies for both architectures.

\vspace{2mm}
\noindent \textbf{Adaptation to domain shift:} We demonstrate the robustness of our adaptive classifier to domain shifts by applying different types of \emph{never-seen-before} distortions with various intensities to the images. One type of distortion is random removal of pixels and replacing them with zero values. This creates salt-and-pepper noise effect. To simulate the scenario where the test images have different resolution compared to train image, we apply Gaussian blurring on test images. We allow ResNet-MANN to use only 10 samples from each new domain for adaptation. Figure \ref{mann_noise} compares the accuracy of the adaptive classifier against that of AlexNet and ResNet. The adaptive classifier outperforms deep networks under all kinds and intensities of noises. As the noise level increases, ResNet and AlexNet accuracies dramatically reduce. In contrast, the classification accuracy of our adaptive classifier remains above 80\% even when accuracies of two popular deep networks' reduce to chance. This experiment indicates that the adaptive classifier is highly robust against changes in data distribution.

\section{Automatic Discovery  of Noisy Labels}
\label{noise_discovery}
The diagnosis error for lung cancer screening is around 30\%, even for experienced radiologists \cite{garland1949scientific}. As a result, the ground truth annotation of lung nodule is imperfect. In section~\ref{deepnet_vs_radiologist}, we have seen that radiologists diagnosis decision is highly variable. Label noise is arguably a major factor hindering deep networks from reaching a clinically useful performance. Recently, a number of techniques have been developed for handling label noise \cite{sukhbaatar2014learning, bootkrajang2012label, patrini2016making, vahdat2017toward}. However, this work either uses heuristic loss correction or requires re-training the estimator for new samples. In this section, we propose a strategy for automatically discovering bad labels. Our method can be applied to any new samples without the need of re-training. This is an important step toward making deep learning more friendly to medical research whose data often come with missing or noisy annotation.  

\subsection{Inconsistency Density Function via Recurrent Networks}
We can find out if a sample is mislabeled by estimating the conditional probability $P(y |\mathbf{x}, \mathbf{x}', y')$, where $(\mathbf{x}, y)$ is a sample with correct label, and $(\mathbf{x}', y')$ is a sample whose label is in question. Intuitively, if $y'$ is also correct,  two samples are more consistent than when $y'$ is wrong. Therefore, $P(y |\mathbf{x}, \mathbf{x}', y')$ tends to have higher value when the label is correct than wrong. In practice, performing such estimate is challenging as the probability function faces high-dimensional data $\mathbf{x}$ in presence of only small number of observations. One can of course perform training using a big dataset to improve the estimate. Techniques like Siamese networks \cite{bromley1994signature} can be adopted for this purpose. However, the trained network cannot be used for the samples in the training set as they might be over-fitted, leading to an over-optimistic accuracy. This defies the original purpose which is removing noisy labels from the training dataset. To this end, we propose the use of MANN for estimating the inconsistency between samples. We then show that this measure is effective for removing noisy labels from training data. 

\vspace{2mm}
\noindent \textbf{Training Procedure:} As before, we define episode as a sequence of observations and labels $\{\mathbf{x}_t, y_t\}_{t=1}^T$. We wish to learn a MANN to predict the conditional probability for a $P(y | \mathbf{x}, \{\mathbf{x}_t, y_t\}_{t=1}^{T'})$, where $T' \le T$. As before, we train the system by sequentially feeding the input $\mathbf{x}_t$ and the time-delayed output $y_{t-1}$ to the network, and predict the current label $y_t$.
More importantly, correct labels are randomly shuffled from episode to episode. For example, nodules can be labeled as 1 in one episode 0 in another. Note that labels are consistent within the same episode. This strategy helps prevent the MANN from learning a fixed mapping from samples to their  class labels \cite{santoro2016one}. Consider the scenario where we don't shuffle the label, the network can simply ignore the label information if it can extract label information from $\mathbf{x}$. This is undesirable as the estimate $P(y|\mathbf{x}, \mathbf{x}', y')$ will not change regardless of whether $y'$ is correct or wrong. Label shuffling forces the network to learn dynamical bindings between images' features and the provided labels, making it more responsive to the change in input labels. This is the key property that we use to remove label noise.

\begin{figure}[!t]
\centering
\includegraphics{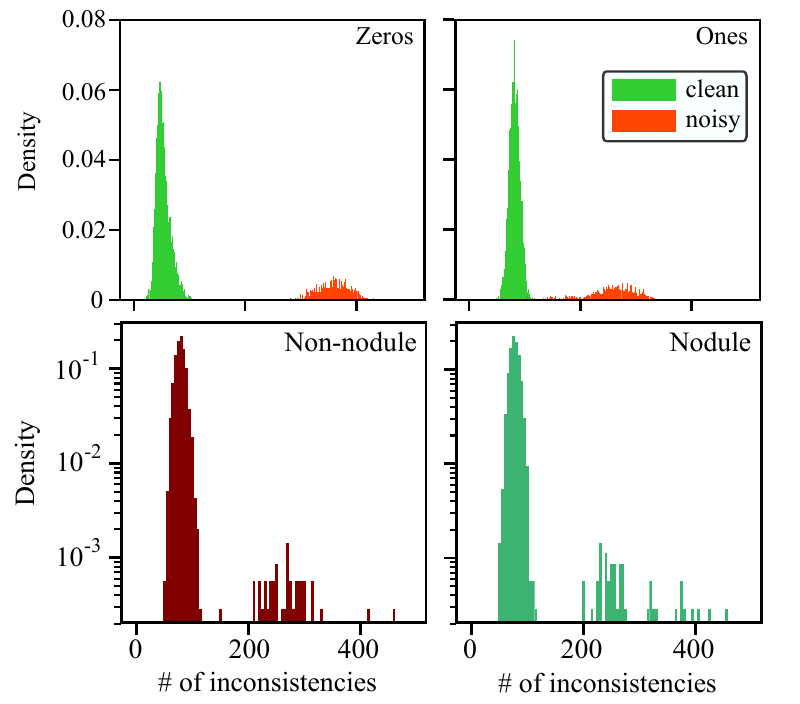}
\caption{IDFs for two classes of MNIST data (top) and lung nodule data (bottom).}
\label{label_noise_idf}
\vspace{-6mm}
\end{figure}

\vspace{2mm}
\noindent \textbf{Discovery of Noisy Labels:} We hypothesize that if we feed MANN with some incorrectly labeled samples $(\mathbf{x}', y')$, then freeze it to perform classification, it will continue assigning incorrect labels to other samples due to its adaptive nature. We call samples that we feed into MANN \emph{seed samples}. Wrongly labeled seed samples will produce higher number of classification errors. In case of lung nodule where the ground truth labels of test samples are uncertain, we use the term \emph{inconsistency} instead of error to indicate a mismatch between classifier's prediction and the provided label. Looking at the statistics related to inconsistency might reveal the subset of mislabeled data. 

As a proof of concept, we first validate the effectiveness of this approach using MNIST dataset, whose true annotation is certain. We randomly flipped the label of 20\% of samples from each class. We use this noisy set to train MANN under binary classification setting. After training, for each sample $(\mathbf{x}', y')$, we feed it to the trained network and compute the number of wrong predictions on a small set of correctly labeled samples. We then plot a inconsistency density function (IDF), defined as probability density function whose x-axis is the number of inconsistencies resulting from feeding a particular seed sample. Top panels of Figure \ref{label_noise_idf} show the resulted IDFs where two distributions of clean and noisy samples are well separated. Assuming a Gaussian distribution for the distribution of clean samples and at confidence level of $\alpha=0.05$, we are able to recognize more than 95\% of the noisy-labeled samples in total (98.7\% and 91.6\% of mislabeled zeros and ones, respectively). 

\begin{figure}[!t]
\centering
\includegraphics{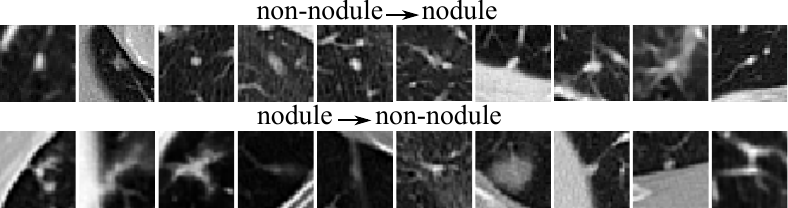}
\caption{Examples of lung nodule images identified by our algorithm as having noisy labels. Labels of these samples were changed during the relabeling.}
\label{relabel2}
\vspace{-6mm}
\end{figure}

We apply the same strategy to lung nodule data to filter out bad labels. The two bottom panels of Figure~\ref{label_noise_idf} show IDFs for both sets of samples labeled as nodules and non-nodules by radiologists. We can observe a similar pattern with MNIST data where IDFs show two distinct modes. One mode with lower mean inconsistency is likely to come from correctly annotated samples, and the other from noisy labels. Since ground truth labels of lung nodule data are uncertain, we request a radiologist to carefully relabel a subset of 54 images belonging to the second mode. Figure~\ref{relabel2} shows examples of these images. 34 out of 54 samples, or 63\%, were assigned different labels by the radiologist. The results from synthetic MNIST experiment and from radiologist's re-examination indicate that the proposed method is highly effective in identifying bad samples within data.

\section{Conclusions}
\label{conclusion}
\vspace{-2mm}
This paper systematically compares performance of deep networks against that of radiology software and radiologists on lung cancer screening task. We found that mistakes made by deep networks are also highly confusing to radiologists. We propose a practical adaptive classifier capable of taking feedback from radiologists and refine its decision. Our experimental results demonstrated that, when data change, the proposed classifier maintains a good performance while popular deep networks' accuracy reduces to chance. Finally, this paper shows that recurrent network can be used to remove noisy label through computing inconsistency density function. The future work will explore different optimization strategies to speed up the convergence of MANN. We will also investigate the efficacy of our label noise removal framework in a general classification setting.

\ifCLASSOPTIONcaptionsoff
  \newpage
\fi



%\bibliographystyle{IEEEtran}


% \end{thebibliography}
\bibliographystyle{IEEEtran}
\bibliography{references}

\begin{thebibliography}{10}
\providecommand{\url}[1]{#1}
\csname url@samestyle\endcsname
\providecommand{\newblock}{\relax}
\providecommand{\bibinfo}[2]{#2}
\providecommand{\BIBentrySTDinterwordspacing}{\spaceskip=0pt\relax}
\providecommand{\BIBentryALTinterwordstretchfactor}{4}
\providecommand{\BIBentryALTinterwordspacing}{\spaceskip=\fontdimen2\font plus
\BIBentryALTinterwordstretchfactor\fontdimen3\font minus
  \fontdimen4\font\relax}
\providecommand{\BIBforeignlanguage}[2]{{%
\expandafter\ifx\csname l@#1\endcsname\relax
\typeout{** WARNING: IEEEtran.bst: No hyphenation pattern has been}%
\typeout{** loaded for the language `#1'. Using the pattern for}%
\typeout{** the default language instead.}%
\else
\language=\csname l@#1\endcsname
\fi
#2}}
\providecommand{\BIBdecl}{\relax}
\BIBdecl

\bibitem{siegel2017cancer}
R.~L. Siegel, K.~D. Miller, and A.~Jemal, ``Cancer statistics, 2017,''
  \emph{CA: a cancer journal for clinicians}, vol.~67, no.~1, pp. 7--30, 2017.

\bibitem{national2011reduced}
N.~L. S. T.~R. Team \emph{et~al.}, ``Reduced lung-cancer mortality with
  low-dose computed tomographic screening,'' \emph{N Engl J Med}, vol. 2011,
  no. 365, pp. 395--409, 2011.

\bibitem{van2009management}
R.~J. van Klaveren, M.~Oudkerk, M.~Prokop, E.~T. Scholten, K.~Nackaerts,
  R.~Vernhout, C.~A. van Iersel, K.~A. van~den Bergh, S.~van't Westeinde,
  C.~van~der Aalst \emph{et~al.}, ``Management of lung nodules detected by
  volume ct scanning,'' \emph{New England Journal of Medicine}, vol. 361,
  no.~23, pp. 2221--2229, 2009.

\bibitem{brady2012discrepancy}
A.~Brady, R.~{\'O}. Laoide, P.~McCarthy, and R.~McDermott, ``Discrepancy and
  error in radiology: concepts, causes and consequences,'' \emph{The Ulster
  medical journal}, vol.~81, no.~1, p.~3, 2012.

\bibitem{brady2016error}
A.~P. Brady, ``Error and discrepancy in radiology: inevitable or avoidable?''
  \emph{Insights into imaging}, pp. 1--12, 2016.

\bibitem{shewaye2016benign}
T.~N. Shewaye and A.~A. Mekonnen, ``Benign-malignant lung nodule classification
  with geometric and appearance histogram features,'' \emph{arXiv preprint
  arXiv:1605.08350}, 2016.

\bibitem{awai2004pulmonary}
K.~Awai, K.~Murao, A.~Ozawa, M.~Komi, H.~Hayakawa, S.~Hori, and Y.~Nishimura,
  ``Pulmonary nodules at chest ct: effect of computer-aided diagnosis on
  radiologists’ detection performance,'' \emph{Radiology}, vol. 230, no.~2,
  pp. 347--352, 2004.

\bibitem{sahiner2009effect}
B.~Sahiner, H.-P. Chan, L.~M. Hadjiiski, P.~N. Cascade, E.~A. Kazerooni, A.~R.
  Chughtai, C.~Poopat, T.~Song, L.~Frank, J.~Stojanovska \emph{et~al.},
  ``Effect of cad on radiologists' detection of lung nodules on thoracic ct
  scans: analysis of an observer performance study by nodule size,''
  \emph{Academic radiology}, vol.~16, no.~12, pp. 1518--1530, 2009.

\bibitem{lee2010random}
S.~L.~A. Lee, A.~Z. Kouzani, and E.~J. Hu, ``Random forest based lung nodule
  classification aided by clustering,'' \emph{Computerized medical imaging and
  graphics}, vol.~34, no.~7, pp. 535--542, 2010.

\bibitem{dou2017multilevel}
Q.~Dou, H.~Chen, L.~Yu, J.~Qin, and P.-A. Heng, ``Multilevel contextual 3-d
  cnns for false positive reduction in pulmonary nodule detection,'' \emph{IEEE
  Transactions on Biomedical Engineering}, vol.~64, no.~7, pp. 1558--1567,
  2017.

\bibitem{firmino2014computer}
M.~Firmino, A.~H. Morais, R.~M. Mendo{\c{c}}a, M.~R. Dantas, H.~R. Hekis, and
  R.~Valentim, ``Computer-aided detection system for lung cancer in computed
  tomography scans: review and future prospects,'' \emph{Biomedical engineering
  online}, vol.~13, no.~1, p.~41, 2014.

\bibitem{shen2017multi}
W.~Shen, M.~Zhou, F.~Yang, D.~Yu, D.~Dong, C.~Yang, Y.~Zang, and J.~Tian,
  ``Multi-crop convolutional neural networks for lung nodule malignancy
  suspiciousness classification,'' \emph{Pattern Recognition}, vol.~61, pp.
  663--673, 2017.

\bibitem{lecun2015deep}
Y.~LeCun, Y.~Bengio, and G.~Hinton, ``Deep learning,'' \emph{Nature}, vol. 521,
  no. 7553, pp. 436--444, 2015.

\bibitem{krizhevsky2012imagenet}
A.~Krizhevsky, I.~Sutskever, and G.~E. Hinton, ``Imagenet classification with
  deep convolutional neural networks,'' in \emph{Advances in neural information
  processing systems}, 2012, pp. 1097--1105.

\bibitem{girshick2014rich}
R.~Girshick, J.~Donahue, T.~Darrell, and J.~Malik, ``Rich feature hierarchies
  for accurate object detection and semantic segmentation,'' in
  \emph{Proceedings of the IEEE conference on computer vision and pattern
  recognition}, 2014, pp. 580--587.

\bibitem{donahue2015long}
J.~Donahue, L.~Anne~Hendricks, S.~Guadarrama, M.~Rohrbach, S.~Venugopalan,
  K.~Saenko, and T.~Darrell, ``Long-term recurrent convolutional networks for
  visual recognition and description,'' in \emph{Proceedings of the IEEE
  conference on computer vision and pattern recognition}, 2015, pp. 2625--2634.

\bibitem{jia2014caffe}
Y.~Jia, E.~Shelhamer, J.~Donahue, S.~Karayev, J.~Long, R.~Girshick,
  S.~Guadarrama, and T.~Darrell, ``Caffe: Convolutional architecture for fast
  feature embedding,'' in \emph{Proceedings of the 22nd ACM international
  conference on Multimedia}.\hskip 1em plus 0.5em minus 0.4em\relax ACM, 2014,
  pp. 675--678.

\bibitem{kalchbrenner2014convolutional}
N.~Kalchbrenner, E.~Grefenstette, and P.~Blunsom, ``A convolutional neural
  network for modelling sentences,'' \emph{arXiv preprint arXiv:1404.2188},
  2014.

\bibitem{graves2013speech}
A.~Graves, A.-r. Mohamed, and G.~Hinton, ``Speech recognition with deep
  recurrent neural networks,'' in \emph{Acoustics, speech and signal processing
  (icassp), 2013 ieee international conference on}.\hskip 1em plus 0.5em minus
  0.4em\relax IEEE, 2013, pp. 6645--6649.

\bibitem{xu2015show}
K.~Xu, J.~Ba, R.~Kiros, K.~Cho, A.~Courville, R.~Salakhudinov, R.~Zemel, and
  Y.~Bengio, ``Show, attend and tell: Neural image caption generation with
  visual attention,'' in \emph{International Conference on Machine Learning},
  2015, pp. 2048--2057.

\bibitem{xu2014deep}
Y.~Xu, T.~Mo, Q.~Feng, P.~Zhong, M.~Lai, I.~Eric, and C.~Chang, ``Deep learning
  of feature representation with multiple instance learning for medical image
  analysis,'' in \emph{Acoustics, Speech and Signal Processing (ICASSP), 2014
  IEEE International Conference on}.\hskip 1em plus 0.5em minus 0.4em\relax
  IEEE, 2014, pp. 1626--1630.

\bibitem{liao2013representation}
S.~Liao, Y.~Gao, A.~Oto, and D.~Shen, ``Representation learning: a unified deep
  learning framework for automatic prostate mr segmentation,'' in
  \emph{International Conference on Medical Image Computing and
  Computer-Assisted Intervention}.\hskip 1em plus 0.5em minus 0.4em\relax
  Springer, 2013, pp. 254--261.

\bibitem{zheng20153d}
Y.~Zheng, D.~Liu, B.~Georgescu, H.~Nguyen, and D.~Comaniciu, ``3d deep learning
  for efficient and robust landmark detection in volumetric data,'' in
  \emph{International Conference on Medical Image Computing and
  Computer-Assisted Intervention}.\hskip 1em plus 0.5em minus 0.4em\relax
  Springer, 2015, pp. 565--572.

\bibitem{cheng2016deep}
X.~Cheng, L.~Zhang, and Y.~Zheng, ``Deep similarity learning for multimodal
  medical images,'' \emph{Computer Methods in Biomechanics and Biomedical
  Engineering: Imaging \& Visualization}, pp. 1--5, 2016.

\bibitem{greenspan2016guest}
H.~Greenspan, B.~van Ginneken, and R.~M. Summers, ``Guest editorial deep
  learning in medical imaging: Overview and future promise of an exciting new
  technique,'' \emph{IEEE Transactions on Medical Imaging}, vol.~35, no.~5, pp.
  1153--1159, 2016.

\bibitem{bar2015chest}
Y.~Bar, I.~Diamant, L.~Wolf, S.~Lieberman, E.~Konen, and H.~Greenspan, ``Chest
  pathology detection using deep learning with non-medical training,'' in
  \emph{Biomedical Imaging (ISBI), 2015 IEEE 12th International Symposium
  on}.\hskip 1em plus 0.5em minus 0.4em\relax IEEE, 2015, pp. 294--297.

\bibitem{roth2014new}
H.~R. Roth, L.~Lu, A.~Seff, K.~M. Cherry, J.~Hoffman, S.~Wang, J.~Liu,
  E.~Turkbey, and R.~M. Summers, ``A new 2.5 d representation for lymph node
  detection using random sets of deep convolutional neural network
  observations,'' in \emph{International Conference on Medical Image Computing
  and Computer-Assisted Intervention}.\hskip 1em plus 0.5em minus 0.4em\relax
  Springer, 2014, pp. 520--527.

\bibitem{roth2015deep}
H.~R. Roth, A.~Farag, L.~Lu, E.~B. Turkbey, and R.~M. Summers, ``Deep
  convolutional networks for pancreas segmentation in ct imaging,'' \emph{arXiv
  preprint arXiv:1504.03967}, 2015.

\bibitem{kumar2015lung}
D.~Kumar, A.~Wong, and D.~A. Clausi, ``Lung nodule classification using deep
  features in ct images,'' in \emph{Computer and Robot Vision (CRV), 2015 12th
  Conference on}.\hskip 1em plus 0.5em minus 0.4em\relax IEEE, 2015, pp.
  133--138.

\bibitem{cheng2016computer}
J.-Z. Cheng, D.~Ni, Y.-H. Chou, J.~Qin, C.-M. Tiu, Y.-C. Chang, C.-S. Huang,
  D.~Shen, and C.-M. Chen, ``Computer-aided diagnosis with deep learning
  architecture: applications to breast lesions in us images and pulmonary
  nodules in ct scans,'' \emph{Scientific reports}, vol.~6, p. 24454, 2016.

\bibitem{mobiny2018fast}
A.~Mobiny and H.~Van~Nguyen, ``Fast capsnet for lung cancer screening,''
  \emph{arXiv preprint arXiv:1806.07416}, 2018.

\bibitem{hua2015computer}
K.-L. Hua, C.-H. Hsu, S.~C. Hidayati, W.-H. Cheng, and Y.-J. Chen,
  ``Computer-aided classification of lung nodules on computed tomography images
  via deep learning technique,'' \emph{OncoTargets and therapy}, vol.~8, 2015.

\bibitem{setio2016pulmonary}
A.~A.~A. Setio, F.~Ciompi, G.~Litjens, P.~Gerke, C.~Jacobs, S.~J. van Riel,
  M.~M.~W. Wille, M.~Naqibullah, C.~I. S{\'a}nchez, and B.~van Ginneken,
  ``Pulmonary nodule detection in ct images: false positive reduction using
  multi-view convolutional networks,'' \emph{IEEE transactions on medical
  imaging}, vol.~35, no.~5, pp. 1160--1169, 2016.

\bibitem{he2016deep}
K.~He, X.~Zhang, S.~Ren, and J.~Sun, ``Deep residual learning for image
  recognition,'' in \emph{Proceedings of the IEEE Conference on Computer Vision
  and Pattern Recognition}, 2016, pp. 770--778.

\bibitem{ioffe2015batch}
S.~Ioffe and C.~Szegedy, ``Batch normalization: Accelerating deep network
  training by reducing internal covariate shift,'' \emph{arXiv preprint
  arXiv:1502.03167}, 2015.

\bibitem{zeiler2014visualizing}
M.~D. Zeiler and R.~Fergus, ``Visualizing and understanding convolutional
  networks,'' in \emph{European conference on computer vision}.\hskip 1em plus
  0.5em minus 0.4em\relax Springer, 2014, pp. 818--833.

\bibitem{kingma2014adam}
D.~Kingma and J.~Ba, ``Adam: A method for stochastic optimization,''
  \emph{arXiv preprint arXiv:1412.6980}, 2014.

\bibitem{storey2003statistical}
J.~D. Storey and R.~Tibshirani, ``Statistical significance for genomewide
  studies,'' \emph{Proceedings of the National Academy of Sciences}, vol. 100,
  no.~16, pp. 9440--9445, 2003.

\bibitem{zhou2014object}
B.~Zhou, A.~Khosla, A.~Lapedriza, A.~Oliva, and A.~Torralba, ``Object detectors
  emerge in deep scene cnns,'' \emph{arXiv preprint arXiv:1412.6856}, 2014.

\bibitem{jorritsma2015improving}
W.~Jorritsma, F.~Cnossen, and P.~van Ooijen, ``Improving the radiologist--cad
  interaction: designing for appropriate trust,'' \emph{Clinical radiology},
  vol.~70, no.~2, pp. 115--122, 2015.

\bibitem{saenko2010adapting}
K.~Saenko, B.~Kulis, M.~Fritz, and T.~Darrell, ``Adapting visual category
  models to new domains,'' \emph{Computer Vision--ECCV 2010}, pp. 213--226,
  2010.

\bibitem{kulis2011you}
B.~Kulis, K.~Saenko, and T.~Darrell, ``What you saw is not what you get: Domain
  adaptation using asymmetric kernel transforms,'' in \emph{Computer Vision and
  Pattern Recognition (CVPR), 2011 IEEE Conference on}.\hskip 1em plus 0.5em
  minus 0.4em\relax IEEE, 2011, pp. 1785--1792.

\bibitem{gopalan2011domain}
R.~Gopalan, R.~Li, and R.~Chellappa, ``Domain adaptation for object
  recognition: An unsupervised approach,'' in \emph{Computer Vision (ICCV),
  2011 IEEE International Conference on}.\hskip 1em plus 0.5em minus
  0.4em\relax IEEE, 2011, pp. 999--1006.

\bibitem{gong2012geodesic}
B.~Gong, Y.~Shi, F.~Sha, and K.~Grauman, ``Geodesic flow kernel for
  unsupervised domain adaptation,'' in \emph{Computer Vision and Pattern
  Recognition (CVPR), 2012 IEEE Conference on}.\hskip 1em plus 0.5em minus
  0.4em\relax IEEE, 2012, pp. 2066--2073.

\bibitem{santoro2016one}
A.~Santoro, S.~Bartunov, M.~Botvinick, D.~Wierstra, and T.~Lillicrap,
  ``One-shot learning with memory-augmented neural networks,'' \emph{arXiv
  preprint arXiv:1605.06065}, 2016.

\bibitem{graves2014neural}
A.~Graves, G.~Wayne, and I.~Danihelka, ``Neural turing machines,'' \emph{arXiv
  preprint arXiv:1410.5401}, 2014.

\bibitem{garland1949scientific}
L.~H. Garland, ``On the scientific evaluation of diagnostic procedures:
  Presidential address thirty-fourth annual meeting of the radiological society
  of north america,'' \emph{Radiology}, vol.~52, no.~3, pp. 309--328, 1949.

\bibitem{sukhbaatar2014learning}
S.~Sukhbaatar and R.~Fergus, ``Learning from noisy labels with deep neural
  networks,'' \emph{arXiv preprint arXiv:1406.2080}, vol.~2, no.~3, p.~4, 2014.

\bibitem{bootkrajang2012label}
J.~Bootkrajang and A.~Kab{\'a}n, ``Label-noise robust logistic regression and
  its applications,'' in \emph{Joint European Conference on Machine Learning
  and Knowledge Discovery in Databases}.\hskip 1em plus 0.5em minus 0.4em\relax
  Springer, 2012, pp. 143--158.

\bibitem{patrini2016making}
G.~Patrini, A.~Rozza, A.~Menon, R.~Nock, and L.~Qu, ``Making neural networks
  robust to label noise: a loss correction approach,'' \emph{arXiv preprint
  arXiv:1609.03683}, 2016.

\bibitem{vahdat2017toward}
A.~Vahdat, ``Toward robustness against label noise in training deep
  discriminative neural networks,'' \emph{arXiv preprint arXiv:1706.00038},
  2017.

\bibitem{bromley1994signature}
J.~Bromley, I.~Guyon, Y.~LeCun, E.~S{\"a}ckinger, and R.~Shah, ``Signature
  verification using a" siamese" time delay neural network,'' in \emph{Advances
  in Neural Information Processing Systems}, 1994, pp. 737--744.

\end{thebibliography}



%\vfill


\end{document}